# Cognitive Residues of Similarity:
## "After-Effects" of Similarity Computations in Visual Search

**Stephanie O'Toole (steph.otoole@gmail.com)**
School of Nursing and Human Sciences, Dublin City University, Glasnevin, Dublin 9, Ireland

**Mark T. Keane (mark.keane@ucd.ie)**
School of Computer Science & Informatics, University College Dublin, Belfield, Dublin 4, Ireland

**Abstract**

What are the "cognitive after-effects" of making a similarity judgement? What, cognitively, is left behind and what effect might these residues have on subsequent processing? In this paper, we probe for such after-effects using a visual search task, performed after a task in which pictures of real-world objects were compared. So, target objects were first presented in a comparison task (e.g., rate the similarity of this object to another) thus, presumably, modifying some of their features before asking people to visually search for the same object in complex scenes (with distractors and camouflaged backgrounds). As visual search is known to be influenced by the features of target objects, then any after-effects of the comparison task should be revealed in subsequent visual searches. Results showed that when people previously rated an object as being high on a scale (e.g., colour similarity or general similarity) then visual search is inhibited (slower RTs and more saccades in eye-tracking) relative to an object being rated as low in the same scale. There was also some evidence that different comparison tasks (e.g., compare on colour or compare on general similarity) have differential effects on visual search.

## Introduction

Although similarity is often touted as a key mechanism in attention, categorisation, thinking and many other cognitive abilities (Medin, Goldstone & Gentner, 1993; Eysenck & Keane, 2010), we actually know very little about the "cognitive after-effects" of similarity judgments; about whether a cognitive residue is left behind after a similarity judgment and, if so, how that residue might influence subsequent processing. Categorisation is perhaps the one obvious exception to this statement. After people perform a categorisation, it is assumed that some conceptual change has occurred; for example, a new abstraction for a collection of concepts may be formed or stored instances may be re-organized[1]. But, in many other cognitive tasks where similarity is implicated, little is said about what changes occur and what is retained in the cognitive system.

Imagine I am walking down the street with a friend and I say "The way the sun is shining today, reminds me of when we were kids going to school in winter". Here, I have made a comparison between a current and past experience that seems too casual to warrant significant categorical change. Yet, a similarity computation has occurred between two experiences; presumably, identifying many specific similarities between their concept-objects and relations (c.f., Falkenhainer, Forbus & Gentner, 1989; Keane, Ledgeway & Duff, 1994). So, what, if anything, of that similarity computation is retained over time? Are all the similarities retained for future reference, in some form, and, if so, for how long?[2] Or are they quickly cast aside as a residue for some cognitive, garbage-collection mechanism? Answers to these questions are important because, if previously computed similarities are available for subsequent processing, then our view of similarity in a whole range of cognitive abilities should radically change.

In this paper, we use a visual search paradigm to probe for the residues of similarity judgments. Successful visual search in cluttered environments (e.g., ones with distracting objects and camouflaged backgrounds) is systematically inhibited by similarities between the searched-for target and aspects of the environment being searched (e.g., Boot, Neider & Kramer, 2009; Neider & Zelinsky, 2005), as revealed by eye-tracking measures. So, in the present paper, to probe for similarity "after-effects", people first performed various comparison tasks and then were asked to search for the objects from these tasks in challenging, visual environments.

## Visual Search

Visual search is fundamental to human cognition. It is a common everyday activity, in which we scan a particular area for an object already known to us (see e.g., Boot, Neider & Kramer, 2009; Neider & Zelinsky, 2005). To find items in our visual field we must continuously shift our visual attention, or "spotlight" (Brefczynski & DeYoe, 1999; Duncan & Humphreys, 1989), from one location to the next navigating an array of distractions. Target detection in visual search is achieved by guiding the fovea to potentially-salient targets in the visual scene using high speed eye movements (saccades; Wolfe et al, 2002). In general, visual search is seen as involving a complex interplay between top-down and bottom-up attention processes (e.g., Mulckhuyse, van Zoest & Theeuwes, 2008; Theeuwes, 2004; Zelinksy, 2008; Wolf et al., 2002).

The top-down aspects of visual search are guided by prior knowledge of the target-object being sought, informed by the features of that object (e.g., its colour or shape) and expectations about where the object is likely to be located (e.g., plug sockets are close to floor level in houses). Such factors are known to improve an observer's speed and accuracy in locating a desired target in a scene (Corbetta & Shulman, 2002). Bottom-up aspects of visual search are often manifested in the way human gaze can be overwhelmingly drawn to salient features of a visual scene; such as a particularly bright colour or an object with vibrant intensity or an object that shares key features with the sought-for target (Corbetta & Shulman, 2002). Specifically, it has been shown that if a number of other objects in the scene share features with the sought-for target (i.e., distractors) or if the background shares features with the target (i.e., it is camouflaged in the background) then search is slower and less accurate (e.g., Neider, Boot & Kramer, 2009). So, as these paradigms show that apsects of the features of targets can affect visual search, they are perhaps good candidates for probing the residues of similarity computations.

[1] Even in this case, it is not wholly clear what is retained of the actual similarity computation; e.g., do we retain all the computed differences and feature similarities found?

[2] For instance, most analogy models implicitly discard their similarity computations, but Gick & Holyoak (1983) argued that analogical induction could occur after several *successful* problem solving episodes with related analogs.

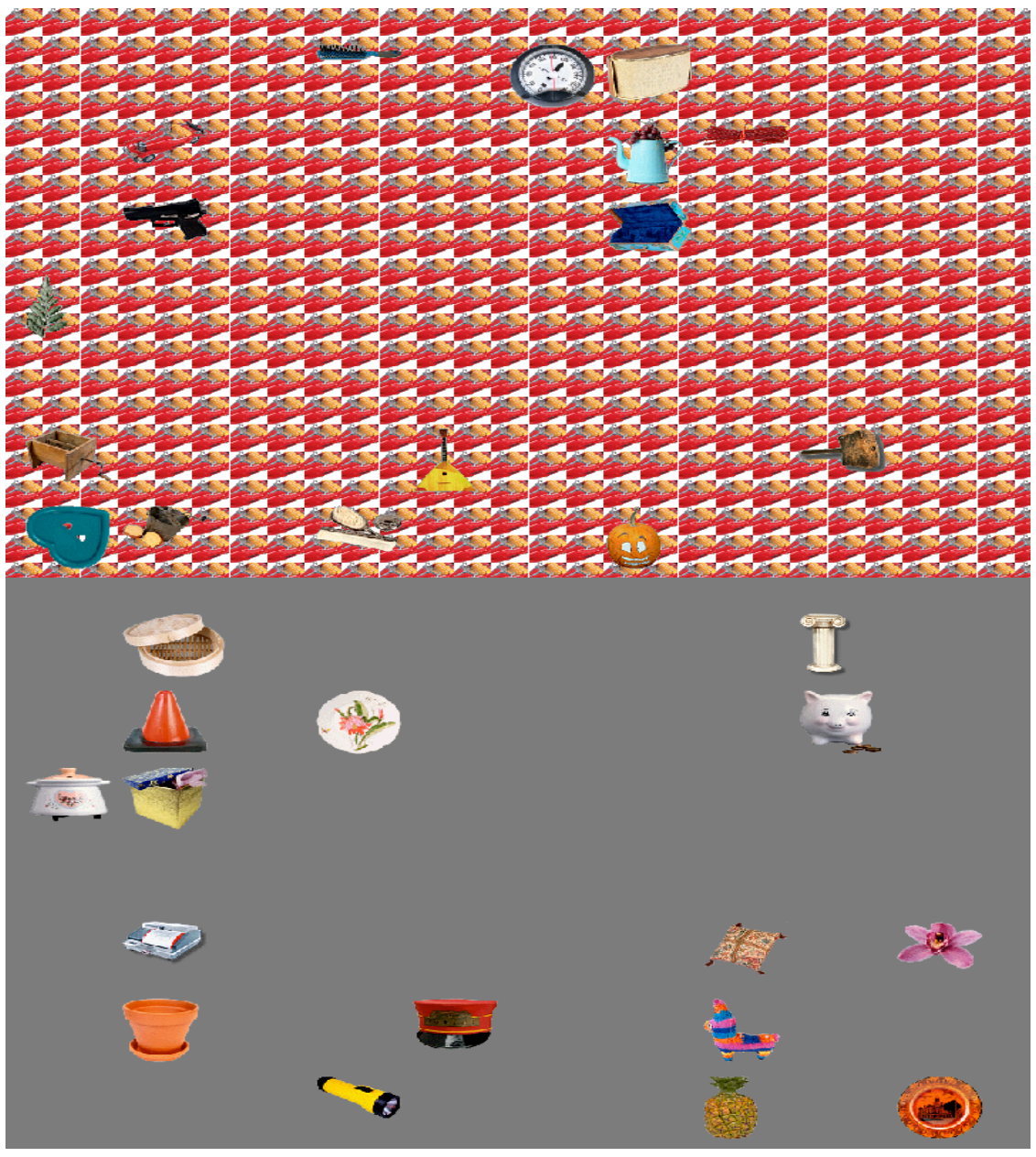

Figure 1: Sample of a camouflaged background where the target object, a red-toy car at the top left, merges with the background and a non-camouflaged background where the target object, a yellow torch, has to be found among distractors.

## Similarity & Visual Search

Real-world environments often provide a continuous background within which to search for stimuli, increasing the complexity of the segmentation that the observer must apply in discriminating an object from its surroundings (e.g., Wolfe et al, 2002). Three key factors have been shown to impact search: target-cue similarity, similarity of and number of distractors to the target and whether the sought-for target is camouflaged relative to the background. All three of these factors are used to increase the difficulty of visual search in our study,

**Target cue similarity** In visual search, target cues are typically manipulated by giving observers a preview of the target object prior to engaging in a search for it. Generally, an observer is asked to view either a relevant target onset or an irrelevant target onset and subsequently asked to search for the target within some context that may or may not include distraction. Mulckhuyse, van Zoest & Theeuwes (2008) found dissimilar target cues yielded shorter participant response times compared to similar target cues. Ludwig & Gilchrist (2003) showed that participants exhibited greater difficulty in disengaging attention from a similar target cue, than from that of a dissimilar target cue.

**Visual similarity of distractor objects** is also known to affect target-object detection in visual search (Neider & Zelinsky, 2005; Zelinsky, 2008). Neider & Zelinsky (2005) found that as the number of distractors increased response times and errors. This work suggests that the presence of visual similarities between the features of objects in the scene and the searched-for target can inhibit successful search. Zhang, Samaras & Zelinsky (2008) also found that when people rated objects as highly similar, they based their judgements largely on the visual attributes of the target, such as colour, texture and shape (see comparison tasks used here).

**Camouflaged backgrounds** created by taking a part of the target object and tiling it as a background, also inhibit visual search (see Figure 1; Boot, Neider & Kramer, 2009). Neider & Zelinsky (2005) showed that camouflaged backgrounds, that maintain high visual similarity to the searched-for target, present very challenging environments in which to search. Neider, Boot & Kramer (2009) manipulated real-world objects as distractor images in camouflaged and non-camouflaged backgrounds (see Figure 1) and found higher levels of inaccuracies and slower response times with high target-similarity backgrounds. However, observers also showed a tendency to direct their gaze to salient distractor images as opposed to the complex background scene, presumably because of the features exhibited by the distractor objects.

## Respects for Similarity

Similarity judgements appear to be straight-forward but they are assuredly not. Medin, Goldstone & Gentner (1993) pointed out the importance of considering the "respects" for similarity; that is, when asked "are these things similar", one immediately needs to answer a further question, namely "similar with respect to what?". Yet, most experiments on similarity simply ask participants to "judge the similarity of these two things" (e.g., Tversky, 1977). Often, if the to-be-compared items are simple geometric shapes the instructions to "judge similarity" is probably interpreted by people as "judge them in terms of their visual features". However, when one moves to real-world objects in real-world environments, asking someone to "judge similarity" could be interpreted as "judge them on their physical appearance" (e.g., they are both pink or round) or "judge them on their functional similarity" (e.g., they can both be used as hammers) or "judge them on the similarity of their environmental contexts" (e.g., you usually find these things together in a kitchen). Hence, in the present study, while some participants are asked simply to "judge the similarity of A and B", others are asked for more specific, comparison tasks.

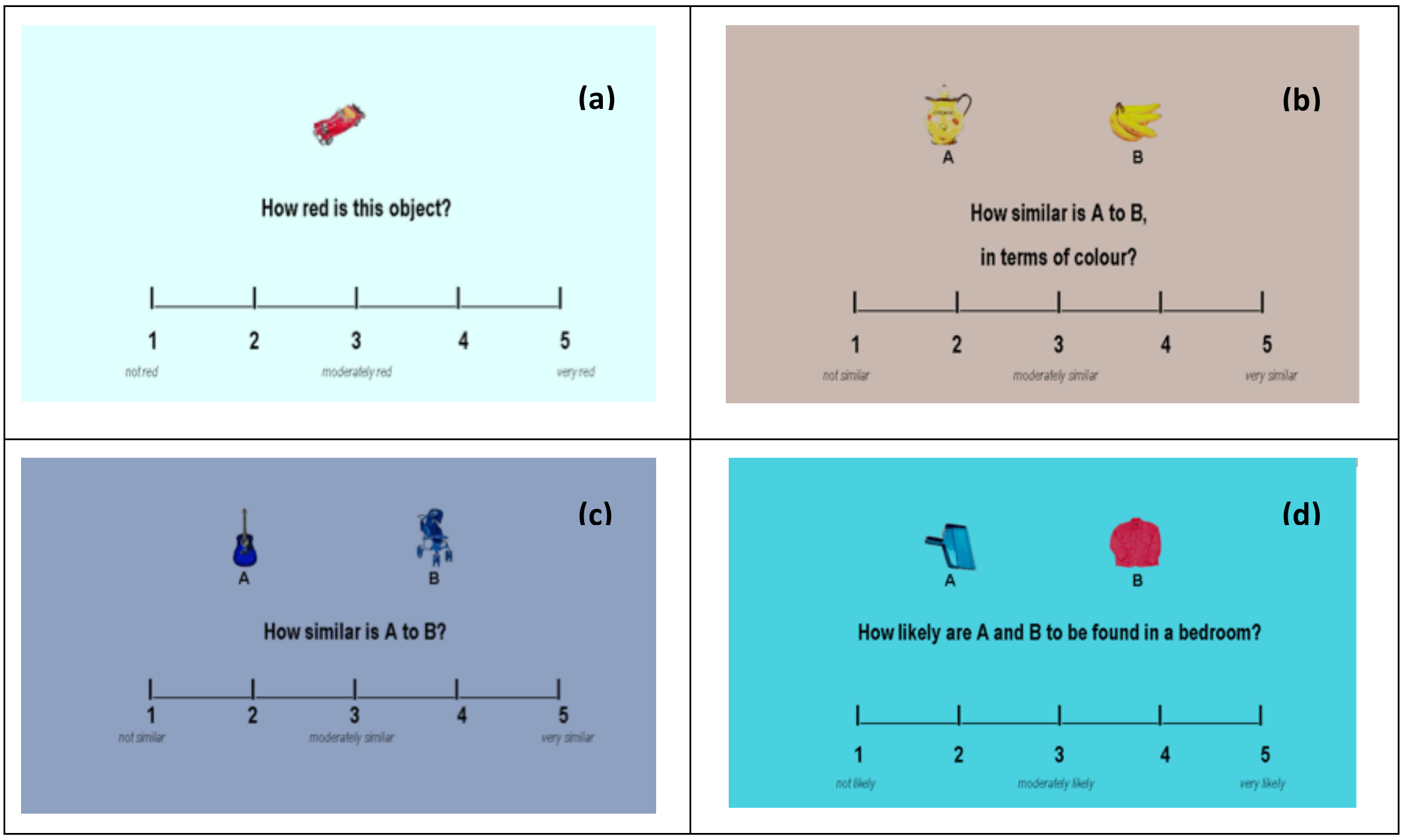

Figure 2: Sample materials from the four comparison tasks:
(a) Single-Colour, (b) Pair-Colour, (c) Pair-General, (d) Pair-Location Task

## The Present Study

The present experiment aims to probe for the residues of similarity computations by seeing whether different comparison contexts affect visual search in complex scenes. So, in the study, every participant performed a comparison task (e.g., similarity rating task) involving 100 trials with various objects before being asked to perform visual searches for these objects. Importantly, the target object used in the visual search was essentially kept constant while aspects of the prior comparison task were varied.

First, participants were assigned to one of four groups, each of which received a different comparison task before doing an identical, visual-search task. In the visual search task, all groups searched for a target object (seen earlier) in a scene containing that object along with 15 distractor objects against camouflaged or non-camouflaged backgrounds (see Figure 1). The four comparison tasks were:

(i) *Colour-Single* received a single, pictured object and were asked to rate the colour of the object on a 5-point scale (see Figure 2a),

(ii) *Colour-Pair* received two pictured objects and had to rate "their similarity to one another in terms of their colour" on a 5-point scale (see Figure 2b),

(iii) *General-Pair* received two pictured objects and had to simply rate their (unspecified) similarity to one another on a 5-point scale (see Figure 2c),

(iv) *Location-Pair* received two pictured objects and had to rate "how likely it was for them to be found" in the same place (e.g., bedroom, kitchen) on a 5-point scale (see Figure 2d),

Note, that two of these comparison tasks involve explicit similarity judgments (i.e., Colour-Pair, General-Pair) while the other two involve, what could be called, implicit similarity judgements (i.e., Colour-Single, Location-Pair). Arguably, the Colour-Single comparison task asks for a comparison between the single, presented object and the category of yellow things in memory and the Location-Pair comparison task essentially asks for a comparison of two objects in terms of where they are typically found. Thus, these manipulations try to explore a number of different "respects" of similarity.

A second, key manipulation was the *rated-context* of the to-be-searched-for object in a given comparison task. Of the 100 items presented (i) 40 were constructed to elicit high ratings (*High-Rated*; i.e., a very red object, two highly similar objects in terms of colour or in general, or two objects highly likely to be found in the same place) and (ii) 60 were constructed to elicit low ratings (*Low-Rated;* i.e., not a very red object, two dissimilar objects in terms of colour or in general, or two objects less likely to be found in the same place). Importantly, this manipulation allows us to see whether specifically finding many similarities *per se* has an effect, independent of the particular comparison task carried out.

Putting all this manipulations together, the experiment had a 4 x 2 x 2 mixed design: 4 (Comparison-Task; Single-Colour, Pair-Colour, Pair-General, or Pair-Location) x 2 (Rating; High versus Low) x 2 (Background; Camouflaged versus Non-Camouflaged); where Comparison-Task was a between-subjects factor and Rating and Background were within-subject factors. In the visual search task, we used an eye-tracker, so our measures were response time and the number of saccades taken to find the target object.

## Method

**Participants** Thirty-two students of University College Dublin (15 females, 17 males) took part voluntarily in the experiment (age ranged from 21 to 27 years, M = 23, SD = 1.12). Eight participants were assigned to each of the four conditions. All participants had normal or corrected-to-normal vision.

**Apparatus** Participants' eye movements were tracked and recorded using a Tobii T60 Eye Tracker. All object images and search displays were presented in full colour on an integrated 17" LCD monitor. The resolution of this monitor was set at 1280 x 1024 pixels and the refresh rate was 60Hz. A 0.5° angle of visual error was allowed after calibration given the availability of free-head motion. Accuracy and response times were recorded via the eye tracker. A keyboard was also used to elicit responses, using the space-bar key. All eye measures (i.e., fixation durations and saccades) were quantified using Tobii gaze plots and heat maps.

**Materials** The experiment had two phases: a comparison task and a visual search task. Pictures of real world objects were used in both tasks. In all, a total of 1600 real world object images (120 target images and 1500 distractor images) were selected from the Hemera Photo Objects Database for use in these materials (Hemera Photo Objects, Gatineau, Quebec, Canada). All were familiar everyday objects such as household items, office supplies and sports equipment. Each object was scaled to fit within an 80 x 80 pixel (3.3° x 3.3°) bounding box.

In the *comparison task*, each the four groups were presented with 120 different materials (first 20 being practice trials) along with a 5-point rating scale and, depending on the group, participants had to rate a single object (Colour-Single condition) or a pair of objects (Pair-Colour, Pair-General, and Pair-Location conditions). The single item or the B-item in the pair was subsequently used as the to-be-searched-for target in the visual search task. Forty of the 100 items were designed to yield a high rating on the scale, in the respective condition, and 60 were designed to yield low rating in the condition.

In the *visual search task*, every group performed 120 visual searches (first 20 being practice trials) using the selected target object from each material in the comparison task. Each target was presented along with 15 distractor objects randomly selected from the 1500 distractor images. In the 100 items, 50 had a camouflaged and 50 a non-camouflaged background.

Following Neider & Zelinsky (2005), all the target and distractor-object stimuli were scaled to fit within a 75 x 75 pixel bounding box. The camouflaged background was created by taking a 20 x 20 pixel square segment from the centre of each object and tiling it continuously across an 800 x 600 pixel blank canvas. That means that each target object had its own unique corresponding background. For the non-camouflage condition a simple and constant gray background was used (see Figure 1). To disperse the distractor stimuli across the visual field, the 800 x 600 pixel canvas was divided into a 10 x 7 grid. To avoid immediate target identification, the target was never presented in the central six locations of the grid. The distractor stimuli were randomly assigned across central and peripheral locations in the grid, with an even amount of stimuli in each of the four quadrants within the grid (see Neider & Zelinsky, 2005; Neider, Boot & Kramer, 2009; Boot, Neider & Kramer, 2009).

**Procedure** The comparison task was always completed before the visual search task. In the *comparison task*, each group received their respective comparison task to complete. Participants were instructed to rate the object/objects on the 5-point scale shown by fixating their gaze directly on the chosen number while simultaneously pressing the space-bar. Once the space-bar was pressed, the next trial would begin immediately. Participants were informed prior to the experiment that they would be assessed both on their response times and accuracy. The order of trials were randomised for each participant.

In the *visual search task*, all participants carried out visual searches for the objects shown earlier. In each trial, they were first presented with a preview of the target object presented on a grey background before being shown the search display, in which the target had to be located amongst distractors against either a camouflaged or non-camouflaged background. After one second the preview was replaced by a search display containing the target object. Participants were instructed to search for the target object and, once found, fixate on the object whilst simultaneously pressing the space-bar. Once the space-bar was pressed, the next trial would begin immediately. All trials were randomly re-ordered for each participant.

## Results & Discussion

The response times and saccades data were collected and analysed for all successful searches. The analysis revealed that the residues of similarity computations affect the ease of successful visual search; when participants had found similarities in the comparison task subsequent visual search was harder. The specific, comparison task also appears to have an effect on visual search.

Separate analyses of variance (ANOVA) were computed for the respond time (RT) and saccade data (Saccades) measures. Eye-tracker data was compiled from the gaze plots of each of the 100 key search scenes during which participants were asked to find an object on the screen. Outlying data points, defined as responses that were 2 standard deviations from a participant's individual mean, were omitted from data analysis.

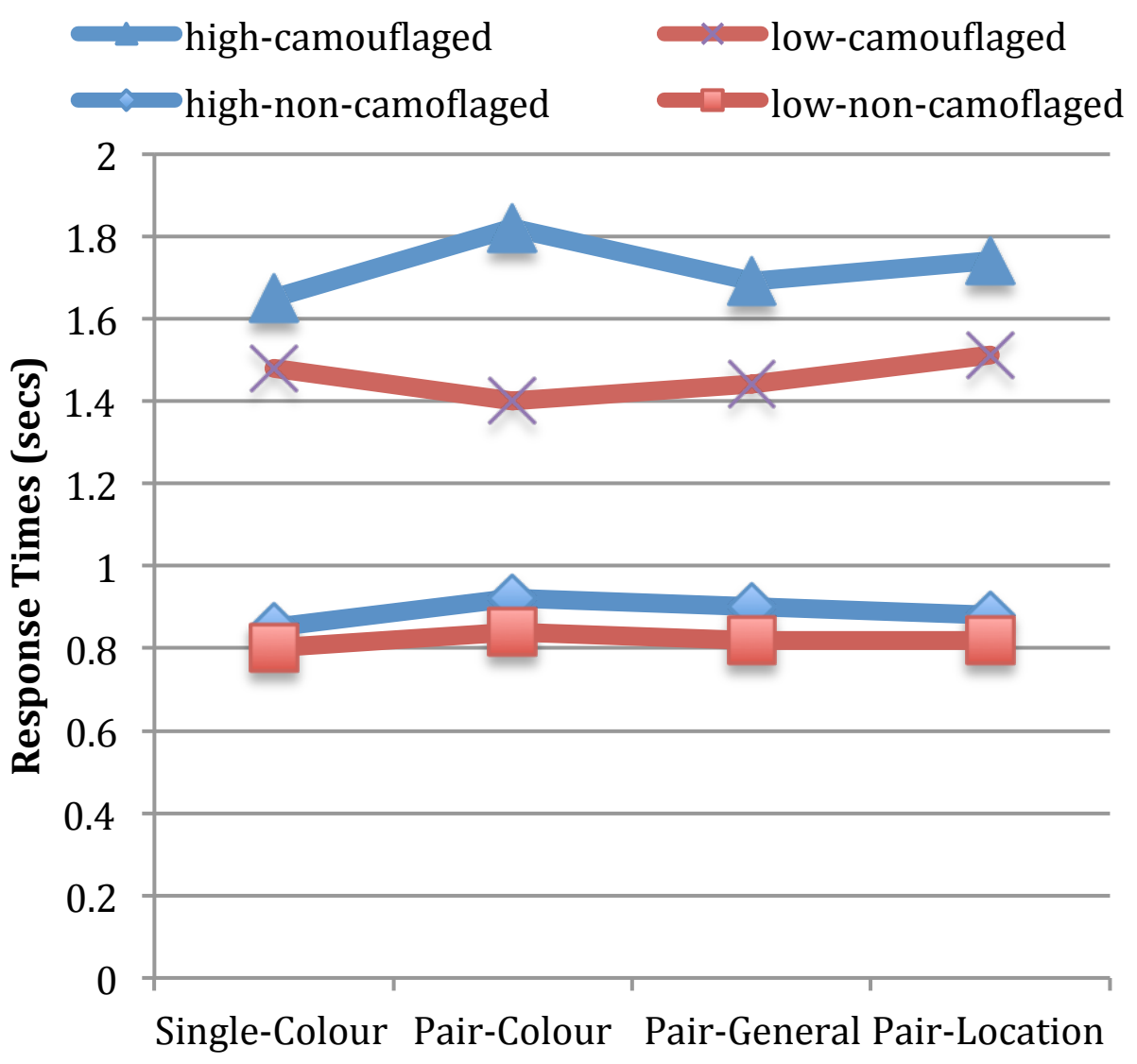

Figure 3: Mean response times in visual search task

### *Response Times (RTs)*

A 4 (Comparison Task) x 2 (Rating) x 2 (Background) ANOVA was conducted for the RT data (see Figure 3). No significant main effect was found for Comparison Task, $F$ (3, .652) = 1.083, $p$ = .647, or Rating, $F$ (1, 1.049) = 2.781, $p$ = .335, though Background was of marginal significance, $F$ (1, .930) = 56.348, $p$ = .09, with camouflaged backgrounds resulting in longer search times than non-camouflaged backgrounds (see Figure 3). Neither was a statistically-reliable interaction found between Comparison-Task and Background, $F$ (3, 3) = .392, $p$ = .769 or Comparison-Task and Rating, $F$ (3, 3) = 1.449, $p$ = .384. However, a reliable interaction was recorded between Rating and Background, $F$ (1, 3) = 17.347, $p$ = .025. The overall three-way interaction was not statistically reliable, $F$ (3, 3022) = .571, $p$ = .634. To tease out these results, we performed two separate analyses of the RT data for search in the camouflaged and non-camouflaged backgrounds.

**RTs for Camouflaged Backgrounds** Analysing RTs for just the camouflaged searches, using a 4 (Comparison-Task) x 2 (Rating) ANOVA, a main effect of Rating is found, $F$ (1, 3) = 26.141, $p$ = .014. However, there was no main effect for Comparison-Task, $F$ (3, 3) = .368, $p$ = .783, and no reliable interaction between Comparison-Task and Rating, $F$ (3, 1497) = .700, $p$ = .552. The main effect found for Rating shows that searches in camouflaged backgrounds are all slower for the High-Rated items relative to the Low-Rated items, suggesting that when similarities are found for the presented target object in the comparison task, subsequent search for this object is hindered in a challenging background. Notably, this effect is found irrespective of the particular comparison task used.

**RTs for Non-Camouflaged Backgrounds** Analysing the RTs for the non-camouflaged scenes, using a 4 (Comparison-Task) x 2 (Rating) ANOVA revealed the same pattern of results found for the camouflaged backgrounds, though there was also evidence of comparison task effects. There were reliable main effects of Comparison Task, $F$ (3, 3) = 9.008, $p$ = .05 and of Rating, $F$ (1, 3) = 88.185, $p$ = .003. There was no reliable interaction between Comparison Task and Rating, $F$ (3, 1525) = .330, $p$ = .803 ($p$ > .05). So, here we again see that the residual similarities formed for High-Rated targets slowed search relative to the Low-Rated targets in which such similarities were absent. Furthermore, different comparison tasks also shows differential search effects.

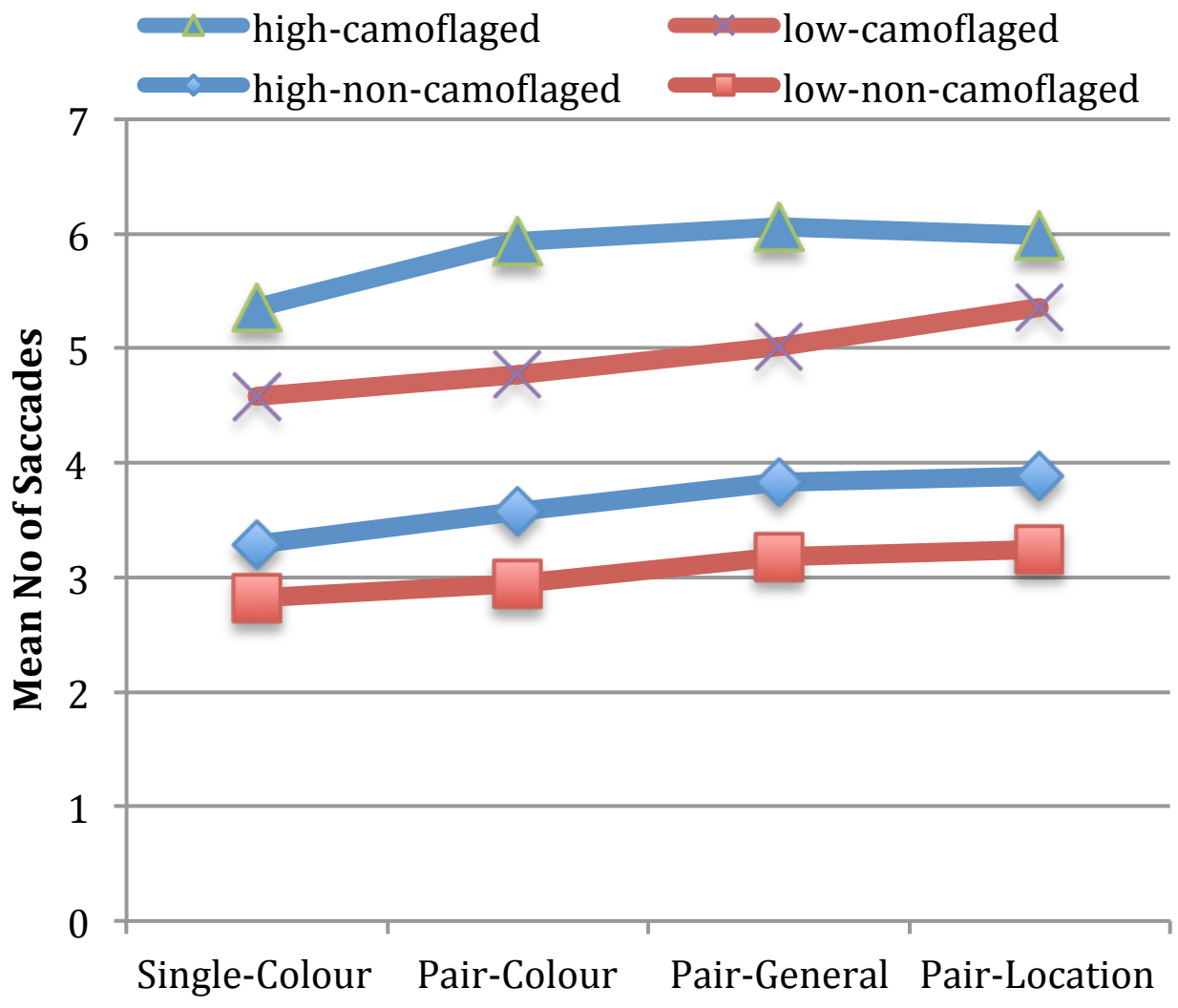

Figure 4: Mean number of saccades in visual search task

### *Saccades*

A 4 (Comparison Task) x 2 (Rating) x 2 (Background) ANOVA was also conducted for the saccade data (see Figure 4). This analysis showed a very similar pattern of results to that found for RTs. No significant main effects were found for Comparison Task, $F$ (3, .923) = 20.294, $p$ = .180, or Rating, $F$ (1, 1.114) = 22.310, $p$ = .114, while Background was marginally significant, $F$ (1, .871) = 189.066, $p$ = .06, with camouflaged backgrounds resulting in longer search times than non-camouflaged backgrounds. No statistically-reliable interaction was found between Comparison-Task and Background, $F$ (3, 3) = .573, $p$ = .670 or Comparison-Task and Rating, $F$ (3, 3) = 1.457, $p$ = .382. However, a marginally significant interaction was recorded between Rating and Background, $F$ (1, 3) = 6.826, $p$ = .07. The overall interaction between all three variables was not reliable, $F$ (3, 3008) = .493, $p$ = .687. Again, to tease out these results, we performed two separate analyses of the saccade data for search in the camouflaged and non-camouflaged scenes.

**Saccades for Camouflaged Backgrounds** Analysing saccades for just the camouflaged searches, using a 4 (Comparison-Task) x 2 (Rating) ANOVA, revealed a main effect of Rating, $F$ (1, 3) = 53.978, $p$ = .005, a marginal main effect for Comparison-Task, $F$ (3, 3) = 6.214, $p$ = .08, and no reliable interaction between Comparison-Task and Rating, $F$ (3, 1491) = .570, $p$ = .635. Again the main effect found for the Rating variable shows that searches in camouflaged backgrounds elicit more saccades for the High-Rated items than for the Low-rated items, suggesting that that when similarities are found for the presented target object in the comparison task, subsequent search for this object is hindered in a challenging background. Though the main effect for Comparison-Task was marginal, planned pairwise comparisons revealed that the Single-Colour condition was significantly different to the Both-General ($p$ = .015) and the Both-Location conditions ($p$ = .002).

**Saccades for Non-Camouflaged Backgrounds** Analysing the saccades for the non-camouflaged backgrounds, using a 4 (Comparison-Task) x 2 (Rating) ANOVA revealed a similar pattern of results to that found for the camouflaged backgrounds. There was a reliable main effect of Comparison Task, $F$ (3, 3) = 28.658, $p$ = .01, and of Rating, $F$ (1, 3.000) = 186.195, $p$ = .001, and no reliable overall interaction between Comparison Task and Rating, $F$ (3, 1517) = .975, $p$ = .403. LSD Pairwise comparisons showed that the Single-Colour task had significantly fewer saccades than the Pair-Colour task on target searches ($p$ = .002), the Single-Colour and Pair-General tasks differed reliably (p = .000), as did the Single-Colour and Pair-Location tasks, ($p$ = .000).

## Conclusions

So, overall, as you would expect, the non-camouflaged searches were more quickly completed than the camouflaged searches. More interestingly, looking for the residue of similarity, we also found that when similarities were computed in the prior, comparison task (High-rating) search times were slower and required more saccades than when such similarities were not found (Low-Rating)[3]. Furthermore, the specific comparison task also appears to have some independent effect on the ease of visual search; with the implicit, Single-Colour task manifesting less interference with visual search relative to the other tasks involving pairs of objects. In short, there appear to be residua left over from similarity computations, that clearly persist for some time to influence subsequent tasks involving the objects that were previously compared and judged to be similar. This raises a significant question about a range of cognitive abilities that rely on similarity computations and how subsequent processing might be influenced by the cognitive "after effects" of similarity processing.

## Acknowledgments

This work is from a self-funded MSc in Cognitive Science at UCD by the first author.

[3] A further experiment, that cannot be reported here for space reasons, replicates these effects.